\documentclass[10pt,twocolumn,letterpaper]{article}

\PassOptionsToPackage{authoryear,round,numbers}{natbib}%
\usepackage[pagenumbers]{cvpr}    
\bibpunct{(}{)}{;}{a}{}{,}

\usepackage[dvipsnames]{xcolor}

\usepackage{multirow}

\PassOptionsToPackage{colorinlistoftodos,textsize=scriptsize}{todonotes}

\definecolor{cvprblue}{rgb}{0.21,0.49,0.74}
\usepackage[pagebackref,breaklinks,colorlinks,citecolor=blue]{hyperref}
\usepackage{nicefrac}

\usepackage{amsmath, amsthm, amsfonts}
\usepackage{graphicx}
\usepackage{multirow}
\usepackage{booktabs}
\usepackage{subcaption}
\usepackage{footmisc}

\usepackage[accsupp]{axessibility} 
\newtheorem{axiom}{Theorem}

\def\paperID{17090} %
\def\confName{CVPR}
\def\confYear{2024}

\title{Scaling Laws for Data Filtering---\\Data Curation \emph{cannot} be Compute Agnostic}

\author{%
 \textbf{Sachin Goyal$^{*\dagger}$\quad \quad 
 Pratyush Maini$^{*\dagger}$} \\
 \textbf{Zachary C. Lipton$^{\dagger}$ \quad \quad 
 Aditi Raghunathan$^{\dagger}$ \quad \quad 
 J. Zico Kolter$^{\dagger,\ddagger}$} \\ 
  Carnegie Mellon University$^\dagger$ \quad\quad Bosch Center for AI$^\ddagger$\\
  \texttt{\{sachingoyal,pratyushmaini,zlipton,raditi,zkolter\}@cmu.edu}
}

\begin{document}
\newcommand{\effectivedata}{$n_{\text{eff}}$}
\newcommand{\utility}{\mathcal{U}}
\newcommand{\trainset}{\mathcal{S}}
\newcommand{\model}{f}
\newcommand{\datapool}{\mathcal{S}}
\newcommand{\D}{\mathcal{D}}
\newcommand{\C}{\mathcal{C}}
\newcommand{\topk}{\texttt{top-k\%}}
\newcommand{\tmars}{T-MARS{ }}
\newcommand{\ar}[1]{{\color{red} [Aditi: #1]}}

\twocolumn[{%
\renewcommand\twocolumn[1][]{#1}%
\maketitle
    \vspace{-14pt}
    \centering
    \hspace{-10pt}
    \begin{minipage}[t]{0.55\textwidth}
        \centering
        \includegraphics[width=\linewidth]{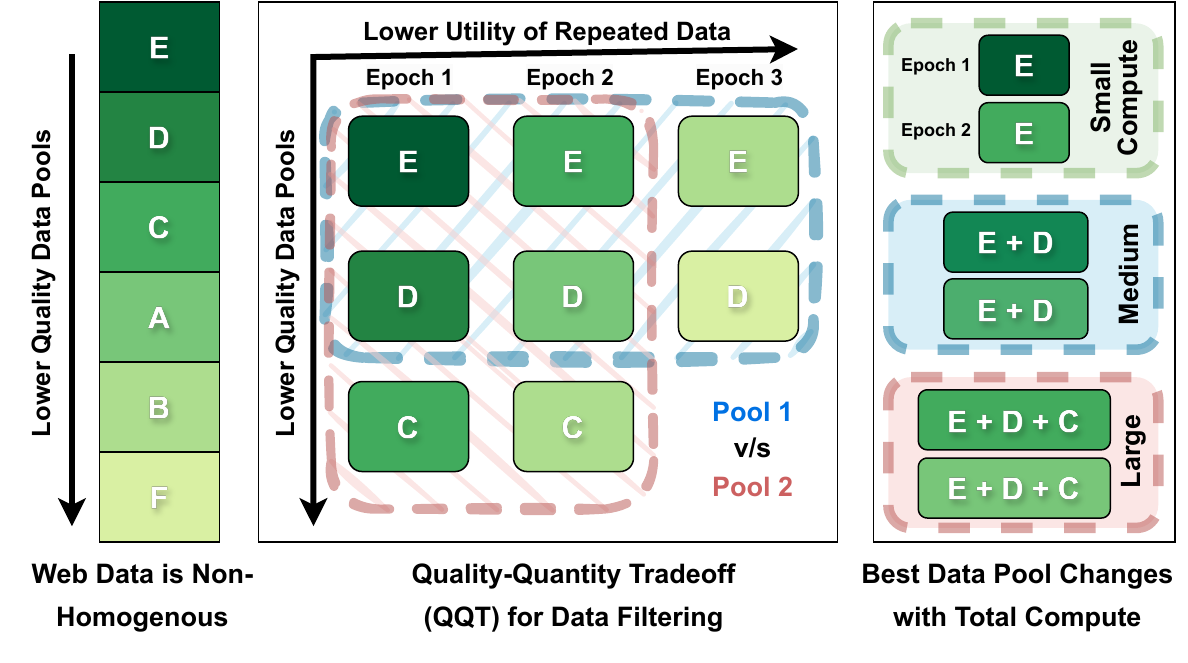}
        \label{fig:task}
    \end{minipage}%
    ~~
     \raisebox{0.30\baselineskip}{%
    \begin{minipage}[t]{0.4\textwidth}
        \centering
        \includegraphics[width=\linewidth]{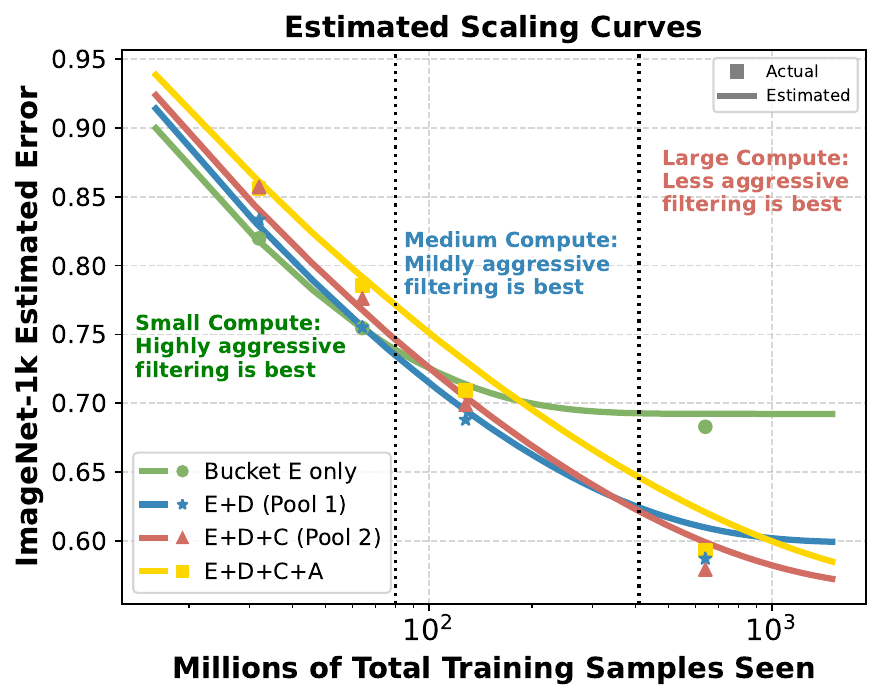}
        \label{fig:tmars_imagenet_scaling_curve}
    \end{minipage}%
    }
    \vspace{-12pt}
    \label{fig:title_figure}
    \captionof{figure}{\textbf{(a)} \textbf{The Dynamic Problem of Data Filtering}: Web data is non-homogenous with subsets of varying quality (y-axis). For pretraining, ``high-quality'' data (such as bucket E) is limited in quantity and loses utility rapidly with repetitions (x-axis), which we call the quality-quantity tradeoff (\texttt{QQT}). Given a fixed compute budget (say equivalent to seeing 6 data pools), should we train on the best pool (E) for 6 epochs or on the 3 best pools (E, D, C) for 2 epochs each (in blue), and so on. 
    How does the answer vary with the total compute budget?  
    \textbf{(b)} We introduce \textbf{scaling laws for data filtering} that accommodate these new axes of \emph{heterogeneous} and \emph{limited} web data. 
    We first model the (differing) initial utility, and rate of decay of utility (scaling parameters)
    of \emph{individual} data pools (such as A--F in (a)). By deriving formulations for the mutual interaction of these buckets, we directly estimate the model performance when trained on combinations of these pools. Importantly, \textbf{our methodology does not involve training on combinations of data pools even for estimating their scaling laws}. 
    Scatter points are true values for comparison, and the solid lines are extrapolated from the scaling parameters of individual buckets.}
    \vspace{16pt}
}]

\renewcommand{\thefootnote}{$^*$}
\footnotetext[1]{Equal Contribution. Code is at {\href{https://github.com/locuslab/scaling_laws_data_filtering}{locuslab/scaling\_laws\_data\_filtering}}}
\renewcommand\thefootnote{\arabic{footnote}}

\begin{abstract}

Vision-language models (VLMs) are trained for thousands of GPU hours on carefully curated web datasets. 
In recent times, data curation has gained prominence with several works developing strategies to retain `high-quality' subsets of `raw' scraped data. 
For instance, the LAION public dataset retained only 10\% of the total crawled data.
However, these strategies are typically developed agnostic of the available compute for training. 
In this paper, we first demonstrate that making filtering decisions independent of training compute is often suboptimal---the limited high-quality data rapidly loses its utility
when repeated, eventually requiring the inclusion of `unseen' but `lower-quality' data.
To address this quality-quantity tradeoff (\texttt{QQT}),
we introduce neural scaling laws that account for the non-homogeneous nature of web data, an angle ignored in existing literature. Our scaling laws
(i) characterize the \emph{differing} `utility' of various quality subsets of web data; (ii) account for how utility diminishes for a data point at its `nth' repetition; and (iii) formulate the mutual interaction of various data pools when combined, enabling the estimation of model performance on a combination of multiple data pools
without ever jointly training on them.
Our key message is that data curation \emph{cannot} be agnostic of the total compute that a model will be trained for. 
Our scaling laws allow us to curate the best possible pool for achieving top performance on Datacomp at various compute budgets, carving out a pareto-frontier for data curation. 
\end{abstract}

\section{Introduction}
\label{sec:intro}

Scaling up foundation model pretraining with more data, compute, and parameters has rewarded the machine learning community with high-performing models. Much of the early research driving the decision 
to `scale up', were governed by the promise of improvement in model performance as estimated by
neural scaling laws~\citep{kaplan2020scaling,chinchilla}. 
In recent times, there has been a general acceptance of `data' being the secret sauce for the best performing `closed models'---be it for LLMs, VLMs, or diffusion models~\citep{achiam2023gpt,dalle3}. With the recognition of the importance of data quality, various streams of work have emerged that have focused on either filtering high-quality data from large corpora~\citep{datacomp,schuhmann2022laion,abbas2023semdedup,rae2021scaling,marion2023less}, or generating new data that is high quality~\citep{nguyen2023improving,li2023textbooks,maini2024rephrasing}. However, past scaling laws were designed considering `data' as one homogenous entity, and never accounted for this important axis of `data quality' that has emerged recently.

\begin{figure}[tbp]
    \centering
    \includegraphics[width=0.5\textwidth]{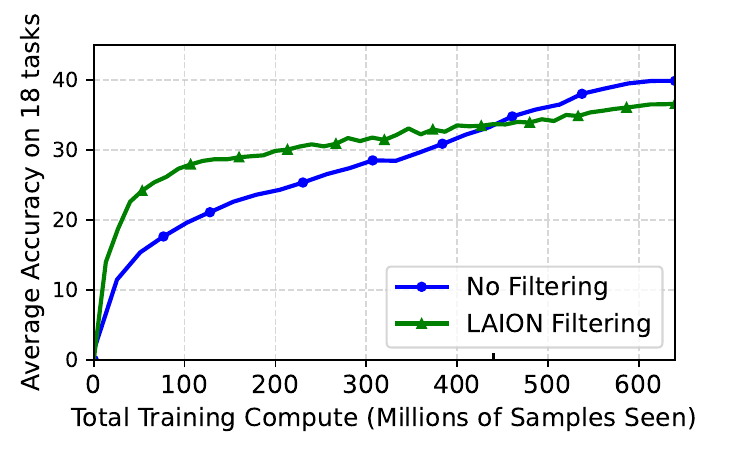}
    \caption{
    Given an initial data pool of 128M samples, we train ViT-B/32 CLIP models for a total of 640M samples.
    As we train for longer, the accuracy gains using the LAION-filtering subset that filters the common crawl to 10\% of its initial size plateau. Surprisingly, even no-filtering of the common crawl is better than the popular LAION-filtering after seeing more than 450M samples.}
    \label{fig:laion_reversal}
\end{figure}

While data on the web is vast, high-quality data (as ascertained by various metrics) is usually limited.
In this work, we first highlight the \textbf{quality-quantity tradeoff} (\texttt{QQT})---a \emph{dynamic} trade-off between training on high-quality data (that is limited), versus training on lower-quality data that is available in large amounts (Figure \textcolor{red}{1}). High-quality data loses its utility when trained on for multiple epochs (as the model has already learned from it). At this point, lower-quality data (which initially had a lower utility) often has higher utility than the repeated higher-quality data. \emph{Under \texttt{QQT}, how do we determine the best data subset to train on?}
To answer this question, any data curation pipeline \emph{must} account for the total compute that the model is trained for. This is in contrast to how the community has considered data filtering; for example, the LAION filtering strategies extract $10\%$ of the highest quality data from the common crawl. But from Figure~\ref{fig:laion_reversal}, it is evident that beyond 35 epochs, training on completely uncurated data is better than training on the LAION-curated\footnote{Note that this refers to a subset of DataComp curated using filtering strategies similar to those of LAION, over and above the default curation used to create the common pool of DataComp. We \emph{do not} use the original LAION dataset itself for training or any other purpose.\label{foot:laion}} high-quality data.
Current neural scaling laws cannot model this dynamic trade-off between quality and quantity.
Furthermore, the field of scaling laws for vision-language models is even more nascent, and most progress has been in language modeling alone.

In this work, we tackle three important limitations of past neural scaling laws by
(i) considering the `quality' axis when scaling up data; (ii) estimating scaling laws for combination of data pools (without ever actually training on the combination), which in turn helps guide optimal data curation decisions; and
(iii) adapting LLM scaling laws to contrastive training settings (like CLIP) where each batch has a squared number of comparisons contributing to loss.

We introduce the \textbf{first scaling laws tailored for \emph{heterogeneous} and \emph{limited} web data}. Large-scale models are trained on a combination of data pools of various quality.
By modeling the aggregate data utility derived from the scaling parameters of individual data pools (such as A-F in Figure \textcolor{red}{1}(a)), we directly estimate the model performance when trained on any combinations of these pools. Importantly, \textbf{our methodology does not involve training on combinations of data pools even to estimate their scaling laws}; instead, we directly estimate their scaling curves from the scaling parameters of the individual constituent pools.
We make some important departures from the past scaling laws to model the repetitions in the contrastive training regime which allows for $O(n^2)$ comparisons. For example, if the training pool size doubles, the number of comparisons contributing to model loss becomes four times.

We formulate the mutual interaction of data from different pools to estimate model performance under various combinations of the data. This guides optimal data curation strategies adaptive to the available compute.
The key message from our work is that \emph{data curation cannot be agnostic of compute}. When training for a low compute budget (less repetitions), quality triumphs under \texttt{QQT}, as highlighted by the best performance of aggressive filtering (Bucket E) under low compute in Figure \textcolor{red}{1}. On the other hand, at computing scales much larger than the available training data, accounting for the diminishing utility of limited high-quality data becomes crucial. This leads to less aggressive filtering i.e. larger quantity of data giving better performance.

Empirically, we show that our scaling laws for heterogeneous web data allow us to predict the pareto-optimal filtering strategy at various compute budgets ranging from 32M to 640M, using the medium scale pool (128M samples) of DataComp~\citep{datacomp}.

\section{Related Work}
\label{sec:related}
\paragraph{Data Filtering}Vision-language models are trained on noisy webscale datasets, making data filtering a crucial precursor. OpenCLIP~\citep{openclip} tried to reproduce the performance of OpenAI's CLIP~\citep{radford2021learning} by curating LAION-400M~\citep{schuhmann2021laion} dataset. However, their performance still lagged that of CLIP, suggesting the importance of DataCuration. Recently, Datacomp~\citep{datacomp} streamlined the efforts in this direction by releasing a well-crafted benchmark challenge for subset selection from common crawl.

Most of the state-of-the-art data curation approaches involve ranking the data using some metric. For example, LAION~\citep{schuhmann2021laion,schuhmann2022laion} uses a CLIP score based filtering (amongst many other rules), where samples with  a image-caption similarity score lower than 0.28 (as assesed by a pretrained CLIP) are filtered out.~\citet{mahmoud2023sieve,nguyen2023improving} propose to use synthetic-captions generated by an image captioning model~\citep{li2023blip2} to rank the data.
Recently, \tmars~\citep{maini2023t} and CAT~\citep{rdk+23} highlighted that a large fraction of images in these webscale datasets lack any learnable
``visual'' features, and have high similarity with the caption only due to text in the images (OCR) matching the caption. T-MARS filters out $50\%$ of the data based on the CLIP similarity scores after masking the text using an OCR detection algorithm. 
Similarly, C-SSFT~\citep{maini2023t} and DFN~\citep{fang2023data} propose filtering out mislabeled samples by assessing the drop in CLIP scores when finetuning a pretrained CLIP on a held-out validation set. Some other works include~\citet{yu2023devil} which uses a mixture of rules and~\citet{xu2023cit} which uses similarity with downstream metadata.

In this work, we highlight why data filtering cannot be agnostic to training compute and how the ordering varies as one changes the training paradigm. In fact, we showcase LAION filtering (used to train state-of-the-art OpenCLIP models ) can even underperform no-filtering or training on the raw common crawl under certain settings.

\paragraph{Scaling Laws in Language Modeling}
One of the most salient trends in recent deep learning research is the observation that neural network performance tends to improve predictably with increases in model size, data size, and computation. In the domain of language modeling, such observations have been systematized into a set of principles known as \textit{scaling laws}.~\citet{kaplan2020scaling} conducted a comprehensive study on scaling laws for neural language models. They observed that, given fixed computational budgets, there exists an optimal model size, training data size, and training time. Interestingly, the triple (model size, data size, batch size) corresponding to the state of the art tends to scale in lockstep, 
reinforcing the intuition that larger models require more data and more computation to be trained effectively. This observation is corroborated by \citet{chinchilla, hernandez2021scaling} who delve deeper into training compute-optimal language models and highlight the importance of balancing computation with model and data sizes.
~\citet{sardana2023chinchillaoptimal} propose modifications to incorporate the inference cost into the scaling laws.~\citet{hutter2021learning,bahri2021explaining} theoretically study neural scaling laws.

Most closely related to our work,~\citet{muennighoff2023scaling} show that training on tokens beyond four epochs yields negligible gains compared to training on new language data due to diminishing utility. 
However, they do not consider the case of different data quality pools. In this work, we how that mixture of data pools cannot be modeled with an effective dataset size formulation of~\citet{muennighoff2023scaling}. Crucially, one needs to model a decay in utility factor (the scaling parameter $b$ in $y=an^b$) as well. 

Finally,~\citet{pmlr-v139-hashimoto21a} study scaling laws for various mixture proportions, but their study is limited to small-scale supervised learning tasks. In this work, we focus on scaling laws for large-scale contrastive training of visual language models like CLIP.

\paragraph{Scaling laws for downstream performance} Although traditionally the scaling laws have focused on modeling the training loss, recent works have started directly modeling the downstream performance~\citep{gadre2024language, isik2024scaling}.~\citet{caballero2023broken,alabdulmohsin2022revisiting} propose some amendments to estimate downstream performance on image classification and machine transalation tasks respectively. 
In this work, we model ImageNet zeroshot accuracy and an average performance over 18 tasks from DataComp~\citep{datacomp} to fit the scaling curves for data filtering.

\paragraph{Scaling Laws in CLIP}
Application of scaling laws to models like CLIP is still an area of active research. As with the scaling laws observed in pure language models, there's an indication that as the model and data sizes for CLIP grow, its performance on downstream vision tasks improves, albeit with diminishing returns \citep{schuhmann2022laion, gadre2023datacomp}.~\citet{cherti2023reproducible} try to fit standard scaling curves similar to~\citet{kaplan2020scaling} on CLIP models of varying size and architecture. However, note that contrary to language models which are rarely trained with more than 3--4 epochs, CLIP training invovles upto 30--40 epochs even at the largest data scale. As we highlight in this work, one needs to model the diminishing gains of data with repeated epochs, in order to accurately estimate scaling curves for visual-language model training.

\section{Data Filtering for a Compute Budget}
\label{sec:teaser}
\subsection{Experimental setup}
\label{sec:task_setup}
We are given a large initial pool of data to train a VLM (which we use synonymously with CLIP) and want to study the effects of data filtering at different compute budgets.

As our base unfiltered pool, we use the ``medium'' scale of the recently data curation benchmark, Datacomp~\citep{datacomp}. The pool contains 128M samples. In Datacomp, the compute budget is fixed to 128M, with the implicit assumption that data filtering methods will continue to obey their respective ordering in performance as we change the compute budget. In this work, we explicitly consider different compute budgets for training steps:$\{32M, 64M, 128M, 640M\}$ and study the performance of data filtering methods. 
Note that filtering to different amounts (for a fixed compute) changes the number of times each training sample is seen. For example, at a compute budget of 128M, each sample in a filtered pool of 12.8M samples would be seen 10 times. 

We assess the performance of our models based on their zero-shot performance across a diverse set of 18 downstream tasks. This includes both (a) classification tasks like ImageNet, ImageNetOOD, CIFAR10, etc., and (b) retrieval tasks like Flickr and MSCOCO. More details about the downstream evaluation tasks can be found in Appendix~\ref{app:downstream_eval_datasets}.

\subsection{When ``good'' data performs worse}
We start with the popular LAION filtering strategy used to obtain the LAION dataset~\citep{schuhmann2021laion,schuhmann2022laion}. This filters for image-caption pairs with a high similarity score ($>0.28$) as measured by OpenAI's CLIP model. When filtering from common crawl, this threshold amounts to retaining just $10\%$ of the original pool.

We first compare training without filtering (i.e. raw common crawl) with training on LAION-filtered subset, at varying compute budgets. Figure~\ref{fig:laion_reversal} shows the average downstream accuracy on 18 tasks (Section~\ref{sec:task_setup}), as the total training iterations (compute) is scaled from 32M to 640M.
We make the following observations: 
\begin{enumerate}
    \item \textbf{Good data is better at low compute budget}: In the regime of low training compute, utilizing high-quality data (for example, via LAION filtering) is beneficial, corroborating the conventional data filtering intuition. For instance, at 128M training iterations, LAION’s approach of filtering surpasses the no-filter strategy significantly, achieving an increase of 7.5\% zero-shot accuracy averaged over 18 tasks.

    \item \textbf{Data filtering hurts at high compute budget}: The advantage offered by data filtering consistently diminishes as we increase our compute budget. Remarkably, beyond 450M iterations, training on the unfiltered common crawl dataset outperforms that on LAION-filtered data.
\end{enumerate}

Why does the same data filtering, which supposedly picks the `best' data, thereby improving performance at low compute, end up hurting performance at high compute? 

LAION-filtering retains around $10\%$ of the data pool, hence at around 450M compute budget, each sample from the LAION-filtered pool is seen around 32 times. The key insight here is that the same sample, as it is seen multiple times over training, offers a diminishing utility each additional time. The LAION-filtered pool has higher initial utility, which does not degrade much at a low compute budget where samples are not repeated too often. However, at a high compute budget, the utility of the LAION-filtered pool diminishes substantially as the samples are repeated multiple times. Eventually, the unfiltered samples, though starting off with a lower utility, end up suffering a smaller drop in utility as they are repeated less often, even outperforming ``high-quality'' LAION-filtered data at some point. 

\paragraph{Remark.} 
In Theorem~\ref{thm:effective_utility} we will later show that the rate of decay of the utility of a pool is influenced by the size of the pool. In particular, because these models are trained with a contrastive objective offering $O(n^2)$ unique comparisons for a dataset of size $n$, changing the pool size by a factor of $k$, actually ends up increasing the total comparisons by $k^2$. This could potentially mean that the reversal point for LAION v/s no filtering happens much later than 40 epochs when the data pool size is increased by a factor of 10.
That said, the insight from this section underscores the need to tailor the filtering approach to the model’s total training compute, challenging existing practices and offering a new direction for optimizing model performance.

We expand upon modeling and estimating this decay of utility in Section~\ref{sec:scaling_curve_fit}, which one can then use to \emph{adaptively} filter the dataset based on the available compute budget.
\subsection{Data filtering must be compute-aware}
\begin{figure}
    \centering
    \includegraphics[width = \linewidth]{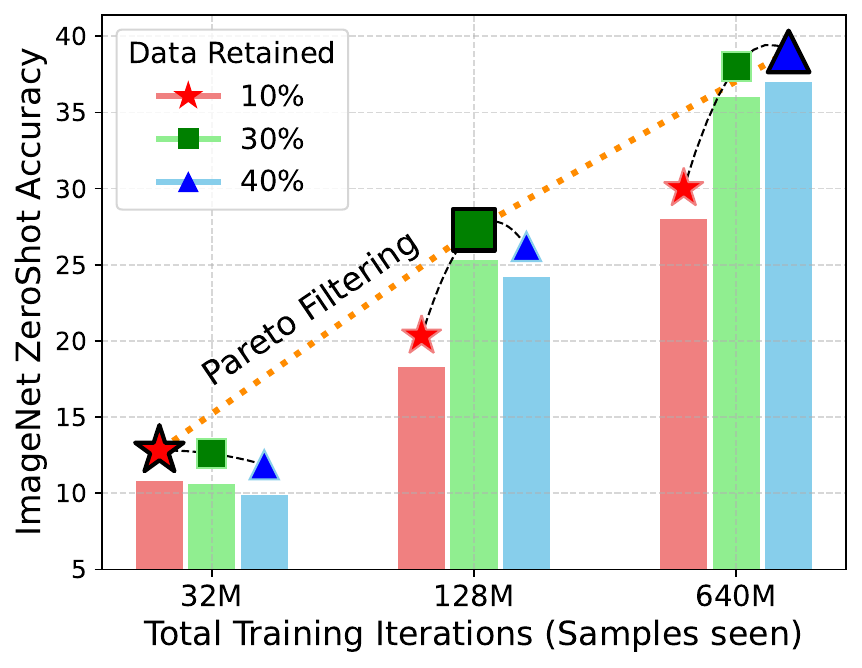}
    \caption{We vary the CLIP filtering threshold after ranking the data by their metric. While the original paper proposed retaining 30\% of the data, our results show that depending on the ratio of compute to data pool size, we must adaptively make the filtering less (or more) aggressive to account for the diminishing utility of good data with repetitions. Results are presented on an average of 18 visual understanding tasks with a global data pool size of 128M samples, and varying compute scales.}
    \label{fig:pareto_clip}
\end{figure}
In the previous section, we saw that the popular LAION-filtering method offered lower gains and eventually under performing the uncurated pool as we increase our training compute. Is this something specific to the LAION-filtering method, or does our intuition about diminishing utility of repeated samples hold for other filtering approaches as well?

We study the performance of some recently proposed state-of-the-art data filtering methods as we change our compute budget. %
We specifically analyze two methods: (a) CLIP score filtering, utilizing the CLIP L/14 model, and (b) \tmars, which ranks data based on CLIP scores after masking text (OCR) features in images (refer to Section~\ref{sec:related}). We compare four levels of varying aggressive filtering  for each data filtering approach, and vary total compute (training iterations) from 32M to 640M, just like before.

Figure~\ref{fig:pareto_clip} illustrates the comparison of Top 10-20\%, Top 30\%, and Top 40\% CLIP filtering at compute scales of 32M, 128M, and 640M. At 32M compute scale, highly aggressive filtering, retaining only the Top 10-20\% data as per CLIP scores, yields the best results, while the least aggressive Top 40\% filtering approach performs the worst. However, this trend \emph{reverses entirely as the compute is scaled to 640M }. While retaining 10\% data excels at low training compute due to fewer repetitions, its utility diminishes rapidly with increased compute due to data repetition.
Similar trends are observed with the \tmars scoring metric (Figure~\ref{fig:pareto_tmars}). Here, retaining Top 20\% data, as originally proposed, stops being optimal as the compute scale increases, and less aggressive approaches prove to be more effective.
These observations underscore the need for a compute-aware filtering strategy that balances the high initial utility of high-quality data which quickly diminishes with repetitions, and the lower-quality but larger data that offers lower initial utility but a slower decay due to fewer repetitions. 

Can we turn this insight into a more performant compute-aware data filtering method? The straightforward strategy is to simply try varying levels of filtering at the compute budget and pick the best. But this is impractical. In Section~\ref{sec:fadu}, we explore
how to \emph{effectively extrapolate} from smaller compute budgets to larger while accounting for diminishing utility with repetition.

\section{Scaling Laws for Data Filtering}
\label{sec:scaling-laws-data}

\subsection{Defining Utility}

Past works on scaling laws~\citep{kaplan2020scaling, jia2021scaling} estimate the error of a model (at a given parameter count) after training for $n$ samples as:
    $y = an^b + d$,
where $a,d>0 \text{ and } b<0$ are constants to be determined empirically, and $y$ is a performance metric such as the loss of the model on a validation set. Intuitively, $b$ factors in
in the diminishing gains as more data is seen and also models the utility of the data pool itself, with a lower value indicating higher utility. For instance, \citet{cherti2023reproducible} noted that the $b$ value for OpenAI's filtered dataset was lower than that of the LAION dataset, indicating it had higher utility.
Whereas, $a$ is a normalizer and $d$ estimates an irreducible error at the end of training to infinity.
Rather than estimating the loss at the end of training for $n$ samples, we can also consider the instantaneous utility of a sample at any point during training. This is given by:
\begin{equation}
\label{eq:instantaneous_utility}
    \frac{dy}{dn} = a\cdot bn^{b-1} = \frac{y}{n}b.
\end{equation}
This equation shows that the instantaneous utility of a sample is proportional to the current loss and inversely proportional to the number of samples seen so far. This is intuitive as the utility of a sample decreases as more data is seen. Crucially, note the data utility parameter $b$. 

\subsection{Utility under Repetition}
\label{sec:main_scaling_law}
Now, let us add one more complexity to this scaling law from past works. In practice, CLIP style pretraining is done for multiple epochs on the same data~\citep{datacomp}. However, there is no clear understanding of how the utility of a sample changes with repetition. We hypothesize that this utility decays exponentially with the number of times the sample is seen. More formally, the utility parameter ($b$) of a sample seen $k+1$ times is given by:
\begin{equation}
\label{eq:utility_repetition}
    b_{k+1} = b \cdot \left(\frac{1}{2}\right)^{\frac{k}{\tau}} = b \cdot \delta^k
\end{equation}
where $\tau$ is the half-life of the utility parameter.  A higher value of $\tau$ indicates that the utility of a sample decays slower with repetition. $\delta$ more concisely captures the decay in utility with repetition, and is used for simplicity of notation.
Then, a closed form expression of the loss of a model after seeing $n$ samples $k$ times each is given by:
\begin{equation}
\label{eq:loss_effective_data}
    y_k =  a\cdot n_1^{b_1}  \prod_{j=2}^k \left(\frac{n_j}{n_{j-1}}\right)^{b_j} + d
\end{equation}
where $n_j$ is the number of samples seen at the end of $j^{th}$ epoch of training. The equation is derived in Appendix~\ref{app:subsec_utility_formulation} and forms the basis of our scaling law.

\paragraph{Summary of Parameters}
Let us concisely summarize the role of each of the parameters in our scaling laws (Eq.~\ref{eq:loss_effective_data}) in order to develop better intuition about each of them.
\begin{enumerate}
    \item \textbf{Utility Parameter $(b)
$}: The change in loss scales with the number of samples seen exponentially based on the value of $b$. A high quality data bucket will have a lower $b$ value (negative with high magnitude) compared to a worse data bucket. 
    \item \textbf{Half life $(\tau)$}: The repetition parameter captures the decay in the utility of repeated data. Intuitively, the half life $\tau$ captures the diversity of the data bucket. Data buckets with high diversity will have a higher value of $\tau$, allowing more repetitions of the bucket, as one would desire. 
    \item \textbf{Decay Parameter $(\delta)$}: The decay parameter is a parameter directly derived from $\tau$, and not a unique parameter. We use this for simplicity of notation. $\delta$ captures the fractional decay in the utility parameter with one epoch of training on that data.
    \item \textbf{Normalizer $(a)$}: The normalizer aims to capture an intrinsic property of the task allowing us to relate the change in loss with the number of samples seen. This does not change with the bucket. We learn a common value of $a$ that minimizes the loss for all buckets, and treat it as a fixed constant across all buckets.
    \item  \textbf{Irreducible loss $(d)$}: This is a constant parameter added to the loss that can not be reduced further.
\end{enumerate}

\subsection{The case of heterogeneous web data}
Now we are ready to add the final layer of complexity to our scaling laws, that of heterogeneous webdata.
A unique challenge in the paradigm of webdata, and critically missed in the existing works on scaling laws, is the presence of data pools of different quality. As discussed in Section~\ref{sec:teaser}, webdata can generally be partitioned into multiple subsets (like using clip score), each  with it's own respective scaling parameters (like respective data utility parameter $b$).

Training large scale models then involves jointly training on a mixture of multiple data buckets. This brings us to us central question---\emph{how can we estimate the loss and thus the scaling curves for a mixture of pools effectively}? This ultimately allows to curate the data conditional to any  compute, rather than a static curation.
One naive way to estimate the error on training on multiple data mixtures would be to use the average error on them. However, this does not factor in the interplay of the two different $b$ values in the exponent of the scaling curve, and how does the repetition parameter ($\tau$) change with increasing data mixtures. 

\begin{axiom}
\label{thm:effective_utility}
Given $p$ data pools \(\mathcal{S}_n^1\) $\hdots$ \(\mathcal{S}_n^p\) sampled uniformly at random, with respective utility and repetition parameters given by $\left(b_1, \tau_1\right) \hdots \left(b_p, \tau_p\right)$, then
the new repetition half-life of each of the buckets $\hat{\tau} = p\cdot \tau$.
Additionally, the effective utility value for the combined pool \(b_{\text{eff}}\)  at the $k^{th}$ repetition is the weighted mean of the individual utility values. Formally, 
\begin{equation}
    b_{\text{eff}}^{(k)} = \frac{\sum_i^p b_i\hat{\delta}_i^k}{p},
\end{equation}
where $\hat{\delta}_i = \left(\frac{1}{2}\right)^{1/\hat{\tau}}$, the new decay parameter per bucket.
\label{theorem:beff}
\end{axiom}

We refer the reader to Appendix~\ref{app:subsec_utility_formulation} for the derivation of the formulae for $b_\text{eff}$.
We assume that the utility of a sample decays exponentially on being seen multiple times. However, there is one major challenge of contrastive training paradigm, where the effective number of samples in a data pool of size $N$ is $N^2$. This is because each sample is paired with every other sample in the data pool. In Appendix~\ref{app:sec:rate_repeat_decay} we show that $\hat{\tau} = \frac{\hat{N}}{N}\tau$ where $\hat{N}$ is the total number of samples in the data pool and $\tau$ is the half-life of the individual pool with \(N\) samples. 

Finally, one can use the $b_{\text{eff}}$ from Theorem~\ref{thm:effective_utility} in Equation~\ref{eq:loss_effective_data}, to estimate the loss while training on a combination of data pools.

\subsection{Various other formulations}
While deciding the scaling laws for data filtering, we considered various other formulations. This included scaling laws that model the decay in `effective samples seen'~\citep{muennighoff2023scaling} rather than effective utility like in this work. In Appendix~\ref{app:scaling_axiom}, we talk in detail about why an ``effective number of samples seen'' based formulation cannot model the heterogeneous webdata and other design considerations. Further, we also study various choices such as the need for allowing different data buckets to have different `half lives' ($\tau$), but a unified normalizer ($a$), and the way we optimized various scaling parameters in Appendix~\ref{app:finding_optimal_scaling_params}.

In \S~\ref{sec:scaling_curve_fit}, we first estimate the scaling parameters for individual data pools of various quality.  Finally, in ~\S~\ref{sec:fadu} we use the scaling parameters of the individual pools to estimate the scaling curves for data combinations.
\section{Fitting scaling curves for various data utility pools}
\label{sec:scaling_curve_fit}

\begin{figure}
    \centering
    \includegraphics[width = \linewidth]{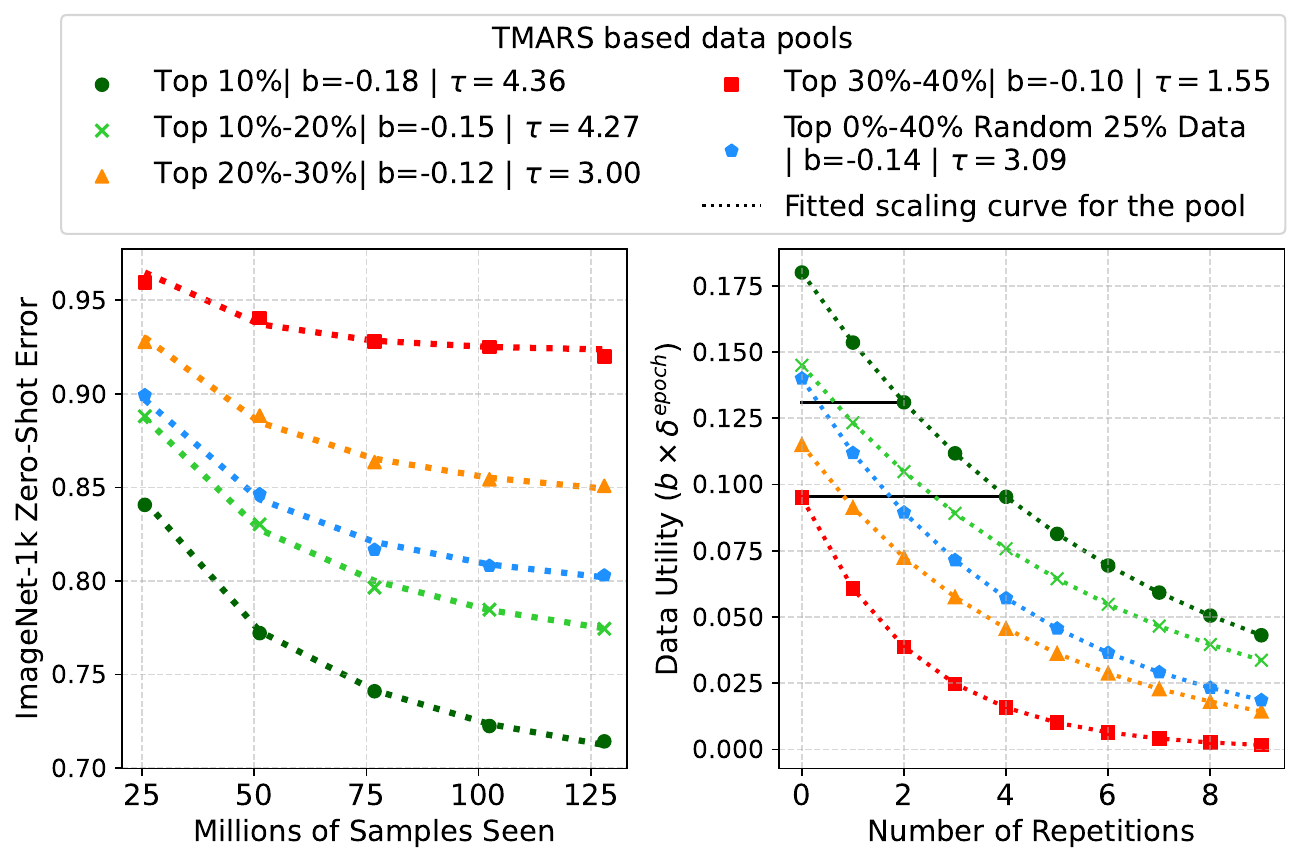}
    \caption{\textbf{Scaling curves for various data quality pools}: We partition the DataComp medium scale pool(128M) samples into various buckets, based on the \tmars scores, and train a model on each bucket for upto 10 epochs. (a) The fitted error curves using the proposed scaling laws (Equation~\ref{eq:loss_effective_data}). (b) Diminishing utilities with epochs of various data subsets. Observe that due to repetitions, even the utility of the best bucket (dark green) at it's $4^{\text{th}}$ repetition becomes lesser than that of worse buckets like top-20\%-30\% (orange curve) subset at it's $1^{\text{st}}$ epoch. This highlights why one needs to adapt the filtering aggressiveness with compute.}
    \label{fig:tmars_scaling_fit}
\end{figure}

\textbf{Experiment Setup: }We experiment on the DataComp medium scale pool which consists of 128M image-caption pairs. In this work, we use T-MARS~\citep{maini2023t} score and CLIP score as the two data utility estimates and rank the web-data based on them. 
Specifically, we form four distinct data subsets, categorized by their respective T-MARS (or CLIP) scores: top 10\% (10\% datapoints with the highest scores), top 10\%-20\%, and so forth, up to the top 30\%-40\% subset. 
Each subset, approximately 12.8M in size, is then used to train a model for 2 to 10 epochs. Finally, we estimate the scaling  parameters (Eq,~\ref{eq:loss_effective_data}), by fitting over the obtained downstream zeroshot error on ImageNet or an average performance over 18 visual classification and retrieval tasks (Appendix~\ref{app:downstream_eval_datasets}). 

Stable optimization of scaling parameters is a crucial step in estimating the scaling laws. This is especially challenging due to the sensitive loss landscape given the complex equations. In this work, we converged at using grid search to estimate the scaling constants $a,b,d \text{ and } \tau$. We detail in Appendix~\ref{app:finding_optimal_scaling_params} on why we made this choice and share the detailed grid used for each of the scaling parameters.

\paragraph{Fitting the scaling laws for individual pools:\;}
Figure~\ref{fig:tmars_scaling_fit} shows  the fitted scaling curves (along with the respective parameters) for various data utility pools using T-MARS score as a data utility metric (See Appendix~\ref{app:scaling_curves_clip} for CLIP score based data pools). The second column in Figure~\ref{fig:tmars_scaling_fit} shows the diminishing utility with epochs of the various data pools. We note some key observations next.

\begin{figure}[h!]
    \centering
    \includegraphics[width = 0.9\linewidth]{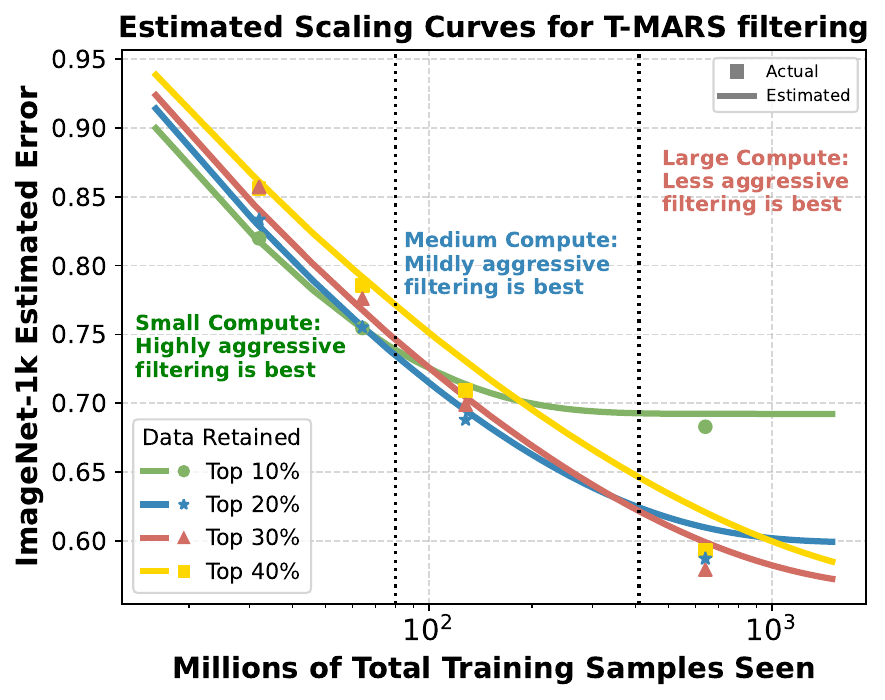}
    \caption{\textbf{Estimating scaling curve for combination of various data pools }: Using the estimated scaling parameters of individual data quality pools (Figure~\ref{fig:tmars_scaling_fit}),
    we estimate the scaling law for the various combinations of the pools by modeling their mutual interaction (Theorem ~\ref{thm:effective_utility}). 
    Note that we do not train on these data combinations to fit the above scaling curves (scatter points are test points), rather the scaling curves are estimated from the scaling parameters of individual pools. Scaling curves for average performance across 18tasks are in Figure~\ref{fig:tmars_18tasks_scaling_curve}.}
    \label{fig:imagenet_tmars_scaling_curve_mixture}
\end{figure}

\paragraph{Web data is heterogeneous and cannot be modeled by a single set of scaling parameters:\;} The heterogeneity of web data is evident from the variability of the data utility parameter $(b)$, which not only varies significantly but also demonstrates a monotonic decrease in magnitude from the highest quality pool (Top 10\%) to the lower quality pool (Top 30\%-Top 40\%). This variation validates the use of $b$ as a metric for data utility. Notably, the overall utility parameter for web data $(b=-0.14)$, depicted by the blue curve, spans the broad spectrum between the highest $(b=-0.18)$ and lowest $(b=-0.10)$ utility parameters. This underscores the inadequacy of a singular scaling law framework in capturing the diverse nature of web data.

\paragraph{Data diversity varies across pools:\;} Figure~\ref{fig:tmars_scaling_fit}  elucidate the variation in the repetition parameter ($\tau$) across the pools, signaling that data diversity is also not uniform. Pools of lower data quality exhibit the smallest values of half-life, indicative of lesser diversity within those pools.

\paragraph{Utility of high quality data with repetitions is worse than that of low quality data:\;} High quality data, despite having a greater initial utility as depicted in the data-quality versus repetitions plot (Figure~\ref{fig:tmars_scaling_fit}, right column), experiences a rapid decline in utility with successive epochs. Notably, the utility of the highest quality data pool (Top 10\%) drops below that of the lowest quality pool (Top 30\%-Top 40\%) after the fourth epoch. This emphasizes that \emph{data filtering will have to be contingent to compute}. 
While training for more compute, including lower quality data in training combination will be benefitial as the limited high quality data will suffer diminishing utility with large number of repetitions.

Finally, it's important to note that this observed diminishing utility is not an artifact of creating subset pools based on T-MARS scores. Similar trends can be seen even with CLIP score based data curation (Appendix~\ref{app:scaling_curves_clip}).

\section{Results: Estimating the Scaling Laws for Data Combinations under QQT}
\label{sec:fadu}

In Section~\ref{sec:scaling_curve_fit}, we derived the respective scaling parameters $a,b,d, \text{and } \tau$ for individual data pools of varying quality. The objective is to determine the most effective data curation strategy given a training compute. By employing Theorem~\ref{theorem:beff} alongside the scaling parameters determined for each individual data pool of varying quality, we now \emph{estimate} the scaling laws for different combinations of these pools. For instance, the Top-20\% pool is considered a combination of the Top-10\% and Top 10\%-20\% data quality pools. The trends from scaling curves can then allow us to predict the pareto optimal data filtering strategy at any given compute.

Figure~\ref{fig:imagenet_tmars_scaling_curve_mixture} and Figure~\ref{fig:imagenet_clip_scaling_curve_mixture} present the scaling curves for different data combinations, evaluating performance on ImageNet. We highlight here that these curves are estimated directly from the scaling parameters of the individual constituent pools, using Theorem~\ref{thm:effective_utility}. We {do not train on the combination of data pools to estimate these scaling curves}. 
The scatter points illustrate actual test performance, serving to validate our estimations. 

\paragraph{Aggressive filtering is best for low compute/less repetitions regime}
Aggressive data filtering proves most advantageous in low-compute regime when repetitions are minimal. This is shown by the superior performance of the highest quality data pool (Top 10\% \tmars score), as illustrated by the green curve in Figure~\ref{fig:imagenet_tmars_scaling_curve_mixture}, when the model is trained for any compute of upto 100M total samples seen. The low compute leads to fewer repetitions, thereby preserving the initial high utility of top-quality data. This trend holds true across both ImageNet zero-shot performance and average performance over 18 tasks (Figure~\ref{fig:tmars_18tasks_scaling_curve} in Appendix~\ref{app:18tasksperf}). Finally, we again note that this is not an artifact for using T-MARS as a metric for data quality. Similar trends can be observed even with CLIP score based data pools, in Figure~\ref{fig:imagenet_clip_scaling_curve_mixture}.

\paragraph{Data curation cannot be agnostic to compute} 
As compute scales beyond 100M samples seen, the optimal data curation strategy shifts. For example, our estimated scaling curves for Imagenet performance for various data pool combinations (Figure~\ref{fig:imagenet_tmars_scaling_curve_mixture}) indicate Top 20\% as the best curation approach when training for 100M to 350M compute (blue curve), rather than the more aggressive filtering of Top 10\% which works the best under 100M training budgets. As the compute scales, the small but high-quality subset of Top10\% suffers from diminishing utility due to lot of repetitions.
On further scaling up the compute beyond 350M samples, even less aggressive filtering strategy of Top 30\% (red curve) works better, highlighting the \texttt{QQT} tradeoff. 
These trends match the pareto optimal strategy as observed empirically as well in Figure~\ref{fig:pareto_tmars}, where at a compute budget of 32M, Top 10\% data retained works the best while at 640M compute, Top30\% works the best.

\subsection{Scaling the scaling curves}
\begin{figure}
    \centering
    \includegraphics[width = 0.8\linewidth]{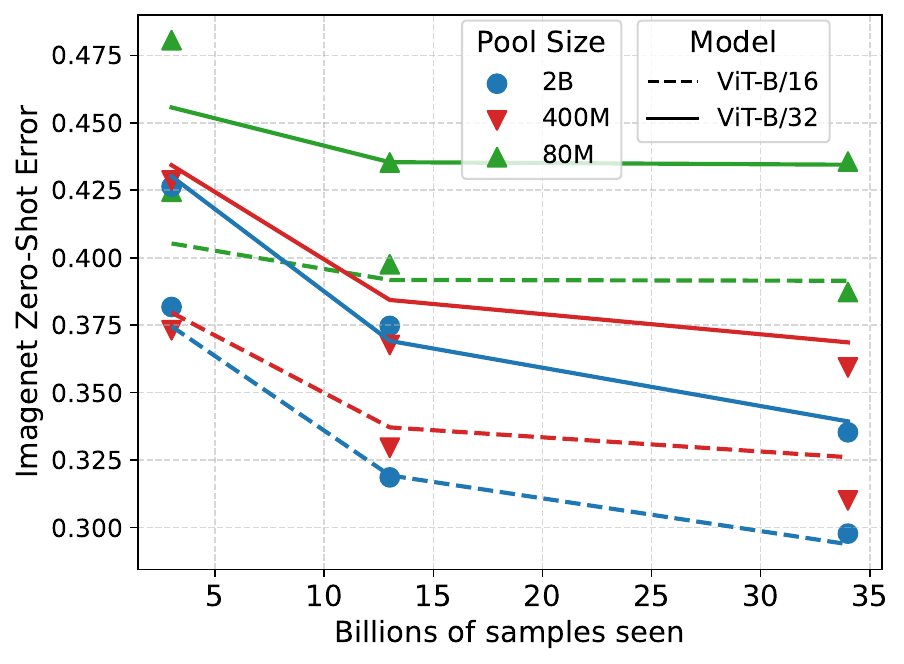}
    \caption{Similar to Figure 5, our scaling law accurately predicts the final error for models trained on 2 different architectures, 3 different very large scale pool sizes and 3 different billion scale compute budgets.}
    \label{fig:laion_scaling_diminishing_util}
\end{figure}

Past work on scaling laws for CLIP models~\citep{cherti2023reproducible} trained tens of models at varying compute scales ranging from 3B to 34B training samples and models spanning different ViT families. While training models at this compute is extremely expensive, we utilize their pretrained models\footnote{Note that these models were trained by ~\citet{cherti2023reproducible} much before the recently pointed out issues in LAION.}. 
~\citet{cherti2023reproducible} tried to fit scaling laws for this family of models, but the scaling curves showed extremely high errors for models trained on small datasets. We believe this is primarily because they do not account for the impact of diminishing utility of repeated data. We use our proposed scaling laws to estimate errors for the models in question. The revised scaling trends are presented in Figure~\ref{fig:laion_scaling_diminishing_util}, which are able to predict the error with a high accuracy. This confirms that our scaling laws hold at massive models trained for 34B data compute, indicating that the diminishing utility of repeated data must indeed be accounted for while predicting model training outcomes.

\section{Discussion}

\paragraph{State of Data Curation}
Despite recent efforts, the curation and utilization of data remains surprisingly ad-hoc and \textit{hacky}, with very little predictability about the outcomes of a filtering strategy. 
In particular, all prior filtering approaches 
(i) propose a metric that ranks examples and filters out data points below a threshold; and (ii) are the thresholds are chosen `agnostic' of the compute the model is supposed to be trained for.
While well-resourced organizations can embark on exhaustive sweeps of `filtering' parameters, this approach (i) is extremely expensive, especially in the paradigm of web-scale pre-training; and (ii) does not transfer to new training paradigms where one changes the training samples to pool size ratios. 

Our scaling laws enable practitioners to precisely assess and quantify the utility of different web data subsets, which is critical given that webdata is heterogeneous. Finally, we show how one can estimate scaling law for a mixture of pool (Theorem~\ref{theorem:beff}). This enables a \emph{compute aware} data curation, where one can decide the filtering threshold (which pools to use for training) based on the estimated accuracies using the scaling law for the mixture of pools.

\paragraph{State of Scaling Laws} To the best of our knowledge, all scaling laws to date have modeled web data with a singular set of scaling parameters, irrespective of the specific formulation of the scaling law. As we venture into the era of large-scale foundation model training, where data curation is a critical step, our work takes significant steps towards estimating the performance of models over various possible choices of combinations of different data quality pools.

\section{Limitations}

\paragraph{Effect of batch-size:\;} Performance of visual language models trained using contrastive loss, varies considerably with the batch size employed during training. Our scaling laws, however, do not account for this variation. We perform all our experiments  with a fixed batch size of $4096$ on the medium scale pool of DataComp. 

\paragraph{Consistency of scaling parameters as the pool size is scaled by orders of magnitude:\;} While we estimate the scaling parameters of different data quality buckets on a given pool size, it is not clear whether the scaling parameters remain same for a similar quality pool of say 100x the size. Crucially, this can allow us to estimate the optimal training subset for a very large scale training by first optimizing the data pools using scaling laws on a smaller scale.

\paragraph{Variation in data diversity i.e. repetition parameter with mixing of pools:\;} In our work, we operate under the assumption that the repetition parameter, influenced by data diversity, remains consistent (up to a factor proportional to the number of mixed pools). Nonetheless, the alteration in diversity across different pools, especially as we blend pools with varying levels of individual diversity, could be more complex or even challenging to predict accurately.

\section{Ethical Considerations}
This study addresses data curation strategies pertinent to web data used in training large-scale foundational models. Such training necessitates meticulous curation to mitigate biases inherent in pretraining datasets. Our methodologies build upon existing strategies such as CLIP filtering and T-MARS filtering, which are designed to be neutral and do not inherently introduce biases. However, all other associated ethical considerations remain applicable. The empirical analyses within this paper were conducted on the standardized DataComp pool, which enforces NSFW content filtering and implements face blurring techniques to uphold privacy standards.

\section*{Acknowledgements}
AR thanks Google for providing GCP credits that supported part of this work. 
In addition, AR thanks
AI2050 program at Schmidt Sciences (Grant \#G2264481).
and Apple for their support.
SG is supported by funding from the Bosch Center for
Artificial Intelligence. 
PM is supported by funding from the DARPA GARD program. 
ZL gratefully acknowledges the NSF (FAI 2040929 and IIS2211955), UPMC, Highmark
Health, Abridge, Ford Research, Mozilla, the PwC Center, Amazon AI, JP Morgan Chase, the Block
Center, the Center for Machine Learning and Health, and the CMU Software Engineering Institute
(SEI) via Department of Defense contract FA8702-15-D-0002, for their generous support of ACMI
Lab’s research.
\clearpage

{
    \small
    \bibliographystyle{plainnat}
    \bibliography{main,paper}
}

\appendix
    \section*{Appendix}
 
    \begin{figure}[ht!]
    \centering
    \includegraphics[width = \linewidth]{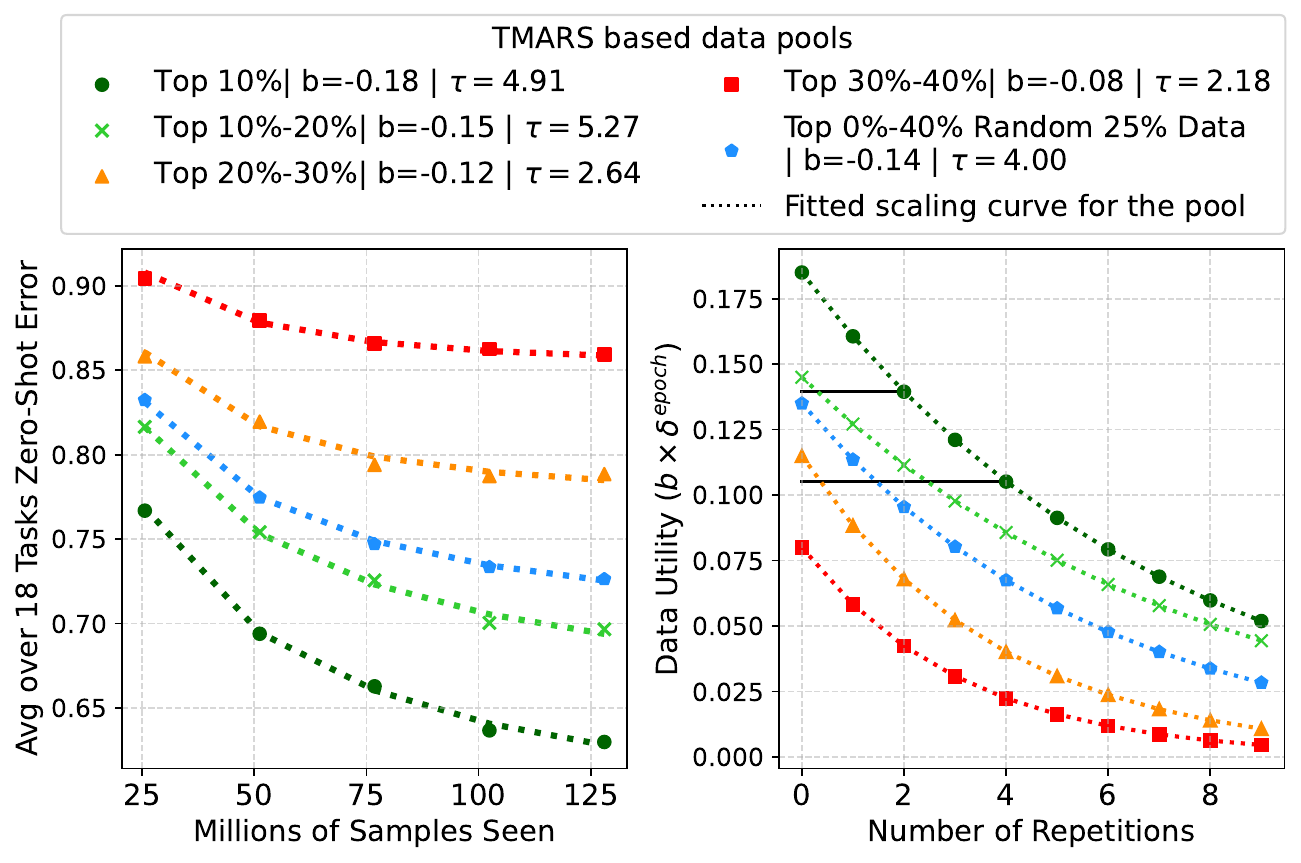}
    \caption{Scaling curves with repeated data for visual-language models: We share the scaling curves for T-MARS based data pools, with average zeroshot performance across 18 tasks as the downstream performance metric.}
    \label{fig:tmars_scaling_fit_18tasks}
\end{figure}

     \section{Scaling curves for average performance across 18 tasks}
     \label{app:18tasksperf}
    In Figure~\ref{fig:tmars_scaling_fit} we showed the obtained scaling curves for each of the data pools, with ImageNet as the downstream performance metric. We share similar curves for each of the data pool, with average performance across 18 tasks as the downstream performance metric, in Figure~\ref{fig:tmars_scaling_fit_18tasks}.

\begin{figure}[h!]
        \centering
        \includegraphics[width = \linewidth]{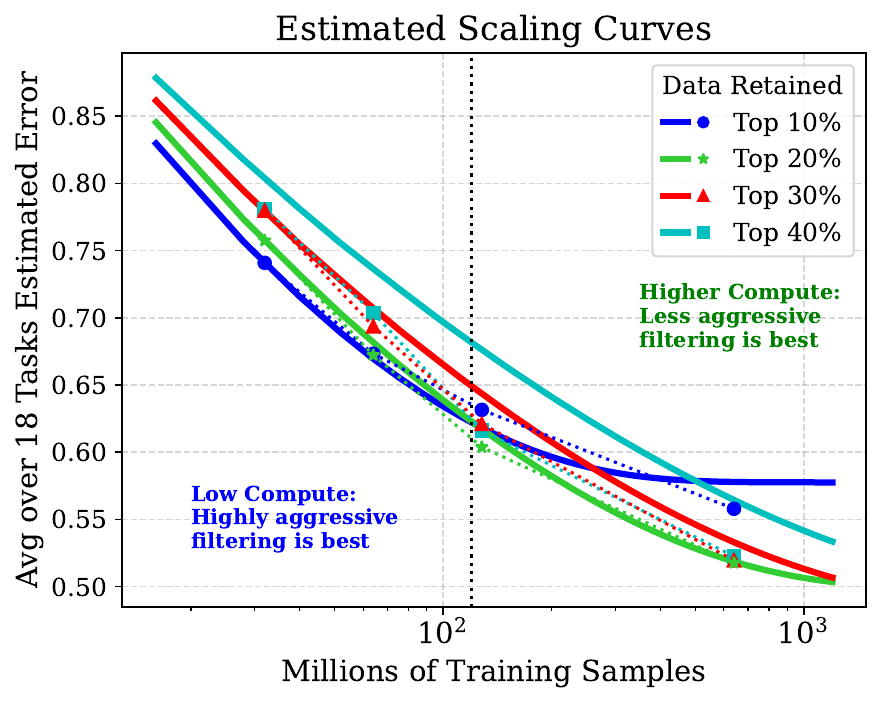}
        \caption{Estimated Scaling curves for different data pool mixtures (data pools based on T-MARS scores). Again, we observe that highly aggressive filtering performs the best at low compute, while moderate filtering is better at higher computes.}
        \label{fig:tmars_18tasks_scaling_curve}
    \end{figure}

    Finally, Figure~\ref{fig:tmars_18tasks_scaling_curve} shows the estimated scaling curves for different data pool mixtures, for average prformance across 18 tasks as the downstream performance metric.

    \section{Downstream Evaluation Datasets and Metrics}
    \label{app:downstream_eval_datasets}
    
    Following prior work~\citep{radford2021clip,wortsman2021robust}, we evaluate our models on a variety of image classification and retrieval datasets to assess their zero-shot capabilities. While the Datacomp~\citep{datacomp} benchmark averages performance across 38 different datasets, we use a subset of 18 such datasets where medium-scale models give better than random performance in order to be able to develop reliable scaling laws. More specifically, we select the following datasets:
    \begin{enumerate}
        \item  ImageNet: a 1000-class image classification challenge~\citep{russakovsky2015imagenet}.
        \item  ImageNet-OOD: Six associated Imagenet distribution shifts---ImageNet-V2~\citep{recht2019doimagenet}, ImageNet-R~\citep{hendrycks2020many}, ImageNet-A~\citep{hendrycks2019natural}, ImageNet-Sketch~\citep{wang2019learningrobust}, ImageNet-O~\citep{hendrycks2019natural}, and ObjectNet~\citep{barbu2019objectnet}. 
        \item  VTAB: 6 out of 12 datasets from the Visual Task Adaptation Benchmark~\citep{vtab}, including Caltech-101~\citep{FeiFei2004LearningGV}, CIFAR10~\citep{Krizhevsky09learningmultiple}, CIFAR100~\citep{Krizhevsky09learningmultiple}, Oxford Flowers-102~\citep{Nilsback08}, Oxford-IIIT Pets~\citep{parkhi12a}, and RESISC45~\citep{Cheng_2017}.
        \item Additional classification datasets: Food-101~\citet{bossard14}, Pascal VOC 2007~\citep{pascal-voc-2012}, and Stanford Cars~\citep{KrauseStarkDengFei-Fei_3DRR2013}.
        \item  Retrieval: 2 retrieval tasks of MSCOCO~\citep{chen2015microsoft} and Flickr~\citep{young-etal-2014-image}.    
    \end{enumerate}
    
    Most of the evaluation datasets constitute image-classification tasks. We use the `Accuracy metric' to evaluate the zero-shot performance of the model on these datasets. The only exceptions include:
    \begin{enumerate}
        \item VTAB: We report `Mean per Class Recall' for Caltech-101~\citep{FeiFei2004LearningGV}, Oxford Flowers-102~\citep{Nilsback08}, Oxford-IIIT Pets~\citep{parkhi12a} datasets. This follows the standard evaluation protocol in past benchmarks~\citep{datacomp} and is done because of the large number of classes in these datasets.
        
        \item Retrieval: For all the retrieval datasets we report the `Mean Recall $@$ 1' which tells how probable is it for the top-recall entry to be relevant.
    \end{enumerate}

    \section{Scaling curves with CLIP Score as the Data Quality Metric}
    \label{app:scaling_curves_clip}
\begin{figure}[h!]
    \centering
    \includegraphics[width = \linewidth]{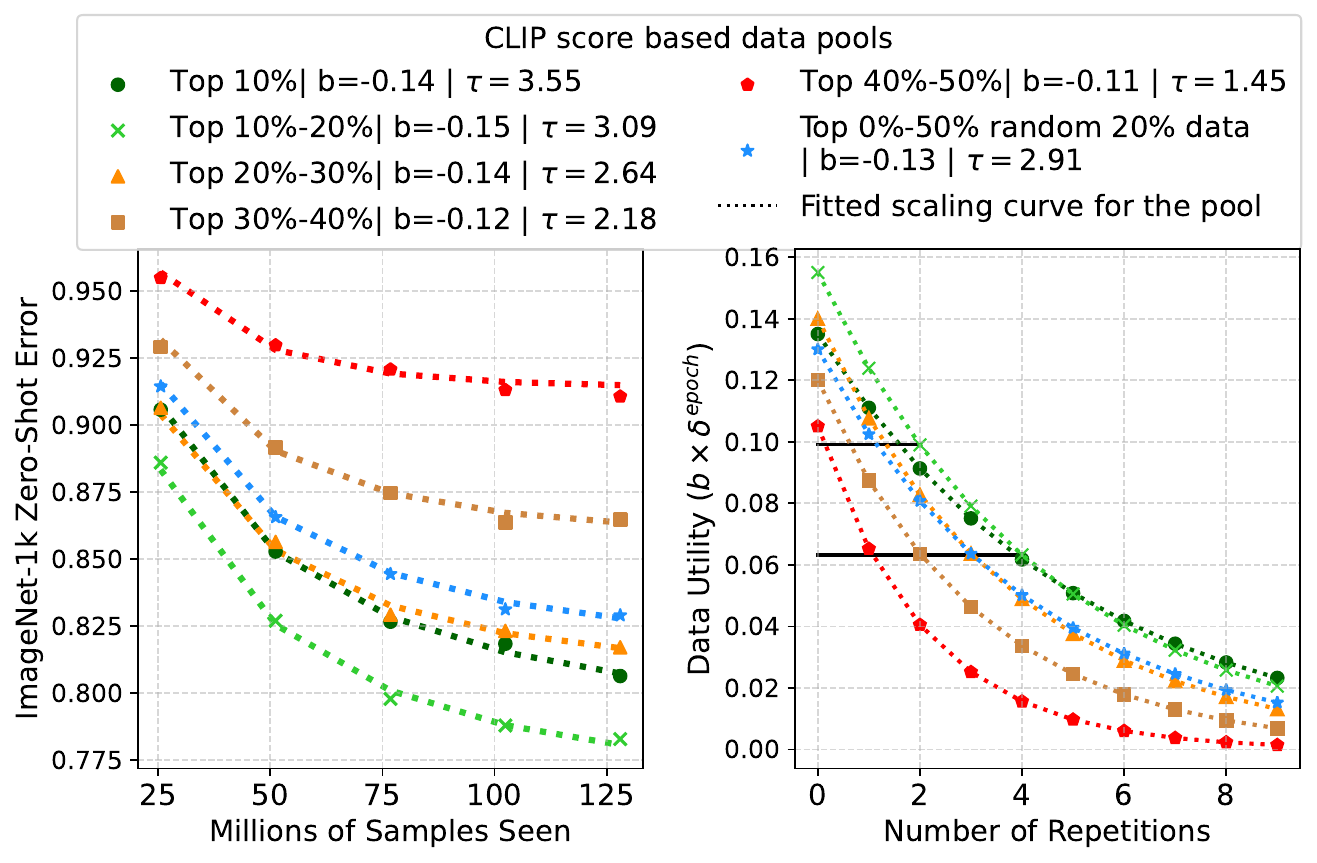}
    \caption{\textbf{Scaling curves for data pools based on CLIP score}: We present the scaling curves for different data quality pools created using CLIP scores. Similar to observations in Figure~\ref{fig:tmars_scaling_fit}, we note that the data quality parameter $b$ decreases in magnitude as we go towards the worse bucket.}
    \label{fig:imagenet_clip_sclaing_fit}
\end{figure}

\begin{figure}[h!]
    \centering
    \includegraphics[width = \linewidth]{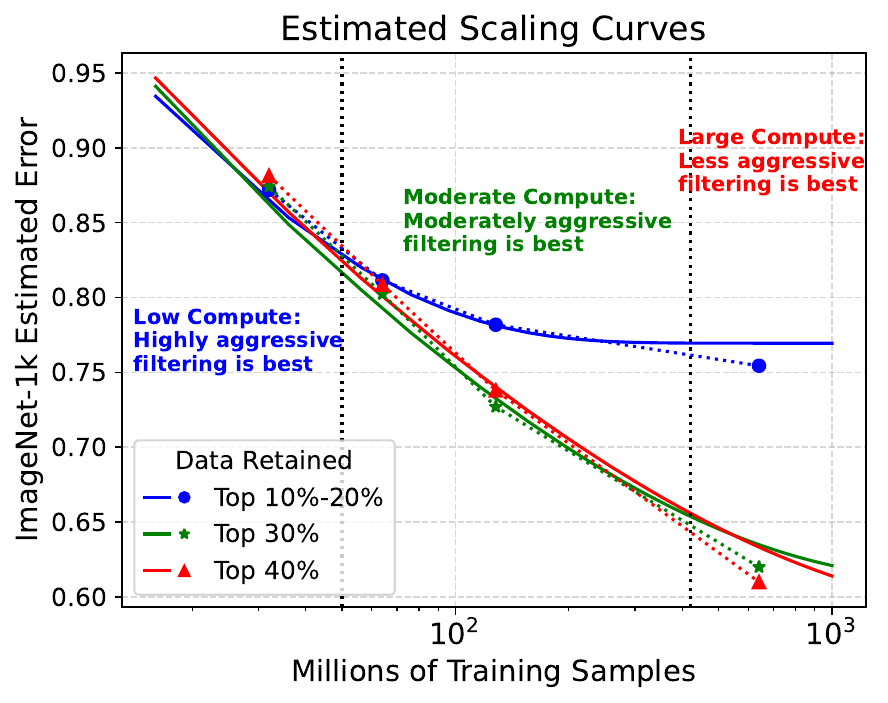}
    \caption{\textbf{Estimating scaling curve for combination of various data quality pools based on CLIP Score}: We estimate the scaling curves for combination of data pools of various qualities, by modeling their mutual interaction (Theorem ~\ref{thm:effective_utility}). Note that we do not train on data combinations to fit the above scaling curves (scatter points are test points), rather the scaling curves are estimated from the scaling parameters of individual pools.}
    \label{fig:imagenet_clip_scaling_curve_mixture}
\end{figure}

    We share the scaling curves for data pools based on the CLIP score in Figure~\ref{fig:imagenet_clip_sclaing_fit}. Similar to the trends observed in \S~\ref{sec:scaling_curve_fit}, we note that the magnitude of data quality parameter $(b)$ decreases as we go towards worse pools. Finally, in Figure~\ref{fig:imagenet_clip_scaling_curve_mixture} we show the scaling curves for combination of pools of varying quality, created using CLIP score filtering.

\section{Pareto data filtering for T-MARS}
\begin{figure}
    \centering
    \includegraphics[width = \linewidth]{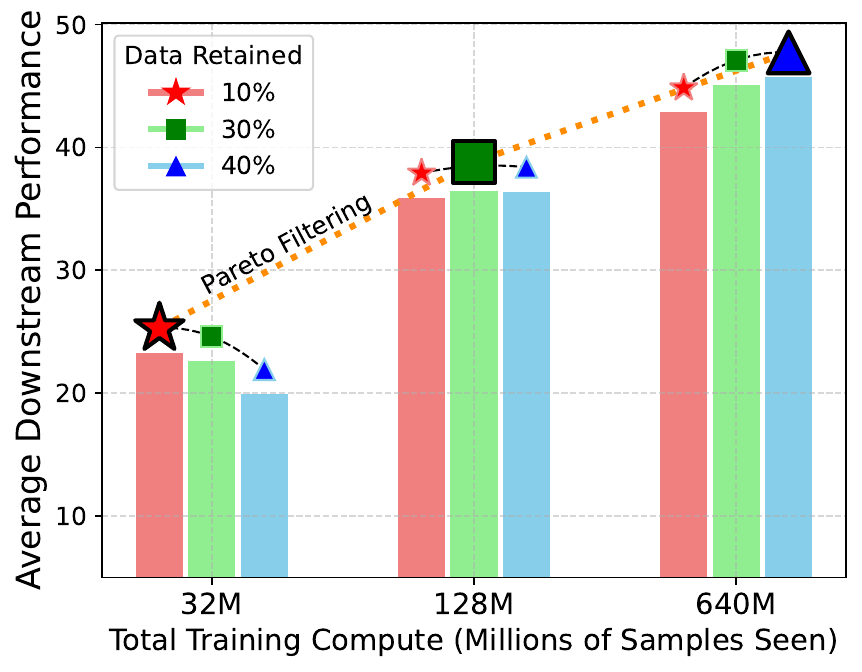}
    \caption{We modify the state of the art data curation approach \tmars by changing the filtering threshold after ranking the data by their metric. While the original paper proposed retaining 30\% of the data, our results show that depending on the ratio of compute to data pool size, we must adaptively make the filtering less (or more) aggressive to account for the diminishing utility of good data with repetitions. Results are presented on an average of 18 visual understanding tasks with a global data pool size of 128M samples, and varying compute scales.}
    \label{fig:pareto_tmars}
\end{figure}

Figure~\ref{fig:pareto_tmars} shows the variation in optimal data filtering strategy for T-MARS as the compute is scaled from 32M to 640M.
    
    \section{Finding the Optimal Scaling Parameters}
    \label{app:finding_optimal_scaling_params}
    Optimizing scaling parameters is a critical step in the validation of any scaling laws, particularly when these models are subjected to scaling laws that govern their behavior across different magnitudes of computation. This process, however, is fraught with challenges, primarily due to the sensitive nature of the optimization landscape, which is characterized by numerous local minima and a complex objective function.
    
    \subsection{Challenges in Parameter Optimization}
    
    In our initial approach to optimizing for the scaling parameters like $a,b,c,\tau$ we employed optimization libraries such as SciPy to find the best fit on the training points. Despite their widespread use, these methods often result in instability and inefficiency in the fitting process, especially being sensitive to initialization. The primary reason for this instability is the high sensitivity of the optimization problem, where slight variations in parameters can lead to significantly different outcomes, and at the same time, the existence of many solutions that fit the curve very well. As a remedy to this, we developed optimization methods ranging from gradient-based optimization to custom-built optimizers. In all of these approaches, we found that we could attain solutions that (a) had very high loss on the data points being optimized over; and (b) conversely, had a low loss on the points it was trying to minimize, but high loss on the points being extrapolated (mixture of data buckets). This was a rather unsatisfactory, because we should find scaling parameters \emph{without} peeking at the points to extrapolate to.
    
    \subsection{Grid Search as a Stable Solution}
    
    Given the limitations of conventional optimization methods, our work advocates for the use of grid search as a more stable and reliable method for determining optimal scaling parameters. Grid search, unlike heuristic or gradient-based optimization methods, systematically explores a specified parameter space, evaluating the performance of each parameter combination to identify the one that yields the best results. This exhaustive search process is especially advantageous in the context of scaling parameter optimization, where the landscape is sensitive to initialization.
    This approach mitigates the risk of overlooking optimal parameters due to the presence of local minima or the complexities of the objective function. Furthermore, grid search facilitates a reproducible parameter fitting process, enabling a clearer understanding of how different parameters influence the model's performance.
    
    \subsection{A view of the loss grid}
    We perform a gridsearch for the normalizer $a\in[0.001,1]$, data utility parameter $b\in[-0.005, -0.5]$, repetition parameter $\tau\in [1,50]$ and $d\in[0.01,0.02,0.05,0.10,0.2]$. These grids were converged upon by continuously expanding the grid limits till the estimated optimal parameters for all the data pools lied in between the grid.

    \subsection{Finding normalizing constant $a$}
\begin{figure}
        \centering
        \includegraphics[width = 0.8\linewidth]{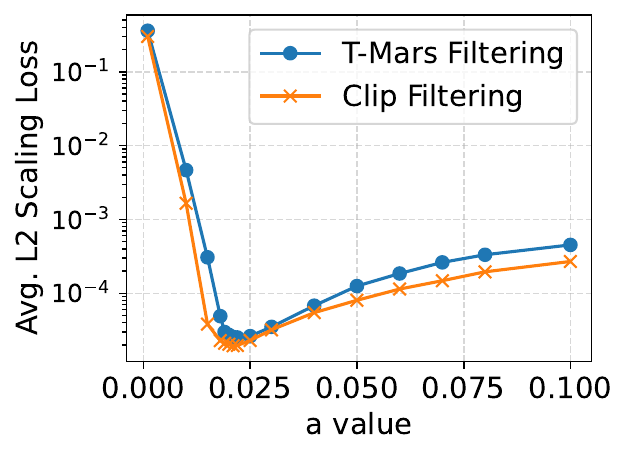}
        \caption{Loss surface as we jointly optimize for $a$ over the data pools. There exists a well-defined minima over our chosen grid-search range for $a$.}
        \label{fig:finding_a}
    \end{figure}

    Recall from \S~\ref{sec:main_scaling_law} that $a$ refers to the normalizer term in the scaling equation which is kept constant across the pools. We learn $a$  by performing a gridsearch over 100 possible values of $a\in [0,1]$, and choosing the one which minimizes the combined loss across all the data quality pools. While we fix $a$ to be same across the pools, we let $b$ and $\tau$ to vary. Figure~\ref{fig:finding_a} shows the loss surface as we jointly optimize for $a$ over data pools. We observe a well behaved loss surface over the gridsearch range, giving an optimal $a$ value.
    
    \subsection{Learning the data repetition parameter $\tau$}
\begin{figure}
        \centering
        \includegraphics[width = 0.8\linewidth]{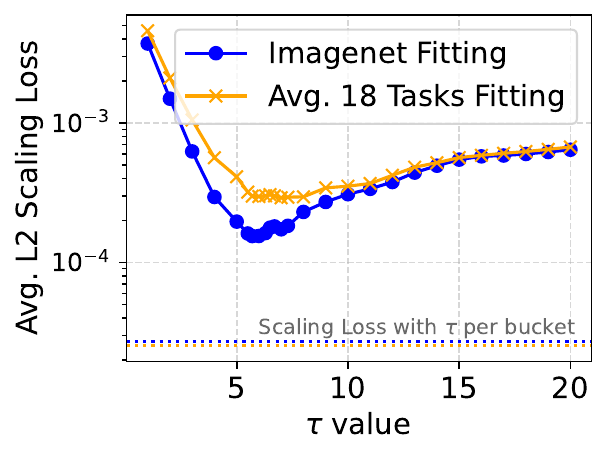}
        \caption{Keeping the data diversity parameter ($\tau$) different across the pools is necessary for good scaling curve fitting. Keeping the $\tau$ value same across the pools give a much worse (higher) L2 fitting loss compared to using different for each pool (see the horizontal dotted line).}
        \label{fig:finding_tau}
    \end{figure}

    Recall from \S~\ref{sec:scaling_curve_fit} that we learnt different data repetition parameter (which denotes data diversity) $\tau$ for each of the data pools. One could alternatively chose to have the same diversity parameter across the pools, assuming that diversity varies uniformly in the web data. However, as shown in Figure~\ref{fig:finding_tau}, learning a same $\tau$ value for all the buckets gives a much worse (high) L2 fitting error compared to learning different values for each of the pool (see the horizontal dotted curve).
    
    \section{What is the Axiom of Scaling?}
    \label{app:scaling_axiom}
    
    Scaling laws have been studied for a while, especially when considering the scaling of large language models with increasing amounts of data~\citep{kaplan2020scaling,chinchilla}. Since these formulations have abstracted away the `per-sample' nature of scaling to the general formulation of the form $y = n^b$, it remains an open question to determine what is the actual `axiom' of scaling. The answer to this question becomes extremely important in enabling us to study the dynamics of a mixture of data buckets of differing utilities, and that too, in an environment where the utility of data is constantly diminishing.
    In this section, we sequentially consider three different `axioms' of scaling. We consider three different formulations that can potentially explain the sample level dynamics of scaling laws---(i) first, we assume that the derivative of the original scaling laws hold universally true, in our utility based formulation; (ii) second we consider the formulation that considers the effective data in the system at any point but considers the utility of data to be constant, as in~\citet{muennighoff2023scaling}; (iii) and finally a formulation that models the decay of data and utility simultaneously. In each of the discussions, we begin with certain assumptions, and examine if those assumptions hold at the event of a mixture of data buckets, that is, starting from an axiom, can we get back to that to predict the system loss in a mixture of data buckets.
    
    \subsection{Utility Based Decay}
    \label{app:subsec_utility_formulation}
    
    We begin with our first scaling law formulation that models only the instantaneous system state and not the global state. The intuition behind such a setup is that it allows the mixing of different buckets naturally by calculating the instantaneous change in the loss by training on any sample from any data bucket. Assume that \ref{eqn:utility_axiom} represents the axiom of scaling of loss with data seen.
    
    \begin{equation}
    \label{eqn:utility_axiom}
    \frac{dy}{dn} = \frac{y(n)}{n}b 
    \end{equation}
    
    The above can be thought of as the derivative of the more generally recognized scaling law $y = n^b$. Then, re-arranging we have that,
    
    \begin{equation}
    \frac{dy}{y} = \frac{dn}{n}b
    \end{equation}
    
    $\Rightarrow$ Assume that `b' utility term is a constant for a given region of the training curve. Then, between two such points in the state where we have seen $n_1$, $n_2$ samples respectively,
    
    \begin{align*}
    \int_{y_1}^{y_2} \frac{dy}{y} &= \int_{n_1}^{n_2} \frac{dn}{n}b \\
    \log \frac{y_2}{y_1} &= b \log n_2 - b \log n_1
    \end{align*}

    \begin{equation}
    \log \frac{y_2}{y_1} = b \log \frac{n_2}{n_1}
    \end{equation}

    Given a single data bucket, the assumption that `b' is constant, should for instance, hold true during a given epoch. This gives us the following closed form relationship between loss values after seeing $n_1, n_2$ samples respectively.
    
    \begin{equation}
    \label{eqn:loss_ratio}
    y_2 = y_1 \left( \frac{n_2}{n_1} \right)^b 
    \end{equation}
    
    The closed-form solution for the rate of decay or loss, $y_k$, between training epochs is given by:
    
    \begin{equation}
    y_k =  y_1 \left(\frac{n_2}{n_1}\right)^{b_2} \left(\frac{n_3}{n_2}\right)^{b_3} \cdots \left(\frac{n_k}{n_{k-1}}\right)^{b_k}
    \end{equation}
    
    Where $y_1$ is the loss at the end of the first epoch, and $$y_k =  n_1^{b_1} \prod_{j=2}^k \left(\frac{n_j}{n_{j-1}}\right)^{b_j}.$$ The scaling term $b$ is assumed to be a constant for a given region of the training curve. For example, during a given epoch, the relationship between $y_2$ and $y_1$ is given by $y_2 = y_1 \left(\frac{n_2}{n_1}\right)^b$.
    
    \subsubsection{Decay of the utility parameter}

    Now, let us move into the paradigm of repeated epochs. We ask the question, how do we model the change in utility of a bucket as we see the dame data point multiple times. 
    We assume that the utility parameter $b$ decays exponentially at every epoch at a constant factor $\delta$. This means that, 
    \begin{align*}
        b^{(k)} = b^{(1)} \delta^{k-1}.
    \end{align*}

    \subsubsection{Estimating Data Mixtures under Utility-based decay}
    Now that we have established the basic laws for the system state under the scaling laws modeling utility-based decay, let us examine how these assumptions play together in a setup of a mixture of buckets. For simplicity, whenever we consider a system with multiple data buckets, we will assume that the data points from these buckets are cycled alternately. Let us assume that we are at a given system state where the model loss and samples seen are determined by $(y_0, n_0)$ respectively. Assume two data buckets parametrized by $(b_1, \delta_1)$ and $(b_2, \delta_2)$ as their respective initial utility, and utility decay rates. Let us first sample a data point from bucket 1, and then from bucket 2. Assume that the loss after seeing a single sample from the two buckets is $y_1, y_2$ respectively. Using the relationship in Equation~\ref{eqn:loss_ratio}:
    
    \begin{align*}
        \frac{y_1}{y_0} &= \left(\frac{n_0 + 1}{n_0}\right)^{b_1} =  \left(1 + \frac{ 1}{n_0}\right)^{b_1} \\
        \frac{y_2}{y_1} &= \left(\frac{n_0 + 2}{n_0 + 1}\right)^{b_2} = \left(1 + \frac{1}{n_0 + 1}\right)^{b_2}
    \end{align*}
    
    Using Taylor expansion and ignoring higher order term assuming that $n_0 \gg 1$, we can write the above as:
    \begin{align*}
        \frac{y_2}{y_0} &= \left(1 + \frac{1}{n_0}\right)^{b_1} \left(1 + \frac{1}{n_0 + 1}\right)^{b_2} \\
        &\approx \left(1 + \frac{1}{n_0}\right)^{b_1} \left(1 + \frac{1}{n_0}\right)^{b_2} \\
        &= \left(1 + \frac{1}{n_0}\right)^{b_1 + b_2}
    \end{align*}
    
    Let us assume that the effective utility of the mixture of data buckets is $b_{\text{eff}}$. Then, in the same 
    steps between $y_0$ and $y_2$, we can write:
    \begin{align*}
        \frac{y_2}{y_0} &= \left(1 + \frac{2}{n_0}\right)^{b_{\text{eff}}}
    \end{align*}
    
    Since the two formulations should be equivalent, we have that
    \begin{align*}
        b_{\text{eff}} &= \frac{b_1 + b_2}{2}
    \end{align*}
    
    This gives us a way to estimate the effective utility of the mixture of data buckets at any given epoch $k$ of training as,
    \begin{align*}
        b_{\text{eff}}^{(k)} = \frac{b_1 \delta_1^{k-1} + b_2 \delta_2^{k-1}}{2}
    \end{align*}

    As we move into the next formulation, let us highlight that the key challenge behind the validity of this formulation is the initial axiom we assumed. We assume that the system is determined by instantaneous interactions where the effective number of samples seen is constant, independent of the number of times it has been seen before, and only its utility decreases every time.
    
    \subsection{Effective Data Based Formulation}
    Let us now consider a formulation that only accounts for the `effective data' in the system to predict the loss of the model on seeing `n' samples. 
    The central assumption behind this formulation is that the `utility' of data itself does not decay, and stays constant throughout the training process, rather only the effective data in the system decays. This is the formulation that has been considered in the past by~\citet{muennighoff2023scaling}.
    Let us consider that the `axiom' of scaling law is now represented by the following equation:
    \begin{equation}
    \begin{split} 
        y &= n_\text{eff}^b,\\
        n_\text{eff} &= \eta(1 + \delta + \delta^2 + \cdots + \delta^{k-2}) + \gamma \delta^{k-1}
        \end{split}
    \end{equation}
    
    where $\eta$ represents the number of unique samples in the training data, $\delta$ represents the rate of decay of effective data when seeing a repeated epoch of the data, and $\gamma < \eta$ represents the number of examples seen in the current epoch of training.
    This suggests that at a given system state, $dn_\text{eff} =\delta^{k-1}dn $
    \begin{align*}
            \frac{dy}{dn_\text{eff}} &= b\frac{y(n_\text{eff})}{n_\text{eff}} \\
            \frac{dn_\text{eff}}{dn} &= \delta^{k-1}
    \end{align*}
    
    Note that since the system state has a fixed utility value throughout training, we can use the formulation from Equation~\ref{eqn:loss_ratio} in terms of $n_\text{eff}$ as follows:
    
    \begin{align*}
        y_2 &= y_1 \left( \frac{n_{\text{eff2}}}{n_{\text{eff1}}} \right)^b
    \end{align*}

    \subsubsection{Estimating Data Mixtures under Effective Data-based decay}
    \label{app:subsubsec_effective_data_formulation}
    Let us now consider the case where we have a mixture of data buckets, and we are trying to estimate loss of the system characterized by the mixture of data buckets. As before, let us assume that we are at a given system state where the model loss and samples seen are determined by $(y_0, n_0)$ respectively. Assume two data buckets parametrized by $(\eta_1, \delta_1)$ and $(\eta_2, \delta_2)$ as their respective initial effective data, and effective data decay rates. Let us first sample a data point from bucket 1, and then from bucket 2. For simplicity, we will consider the case of training on the second epoch of data:
    \begin{align*}
        \frac{y_1}{y_0} &= \left(\frac{n_0 + \delta_1}{n_0}\right)^{b_1} =  \left(1 + \frac{ \delta_1}{n_0}\right)^{b_1} \\
        \frac{y_2}{y_1} &= \left(\frac{n_0 + \delta_1 + \delta_2}{n_0 + \delta_1}\right)^{b_2} = \left(1 + \frac{\delta_2}{n_0 + \delta_1}\right)^{b_2}
        \\
        \frac{y_2}{y_0} &= \left(1 + \frac{\delta_1}{n_0}\right)^{b_1} \left(1 + \frac{\delta_2}{n_0 + \delta_1}\right)^{b_2} \\
        &\approx \left(1 + \frac{\delta_1}{n_0}\right)^{b_1} \left(1 + \frac{\delta_2}{n_0}\right)^{b_2} \\
        &= \left(1 + \frac{\delta_1}{n_0}\right)^{b_1} \left(1 + \frac{\delta_2}{n_0}\right)^{b_2}
    \end{align*}

    By Taylor expansion, and keeping only the first order terms, in the expansion of $(1 + x)^a$ for very small $x$, we have that $(1 + x)^a \approx 1 + ax$. This gives us that,
    \begin{align*}
        \frac{y_2}{y_0} &\approx \left(1 + \frac{\delta_1}{n_0}\right)^{b_1} \left(1 + \frac{\delta_2}{n_0}\right)^{b_2} \\
        &\approx \left(1 + \frac{\delta_1}{n_0}b_1\right) \left(1 + \frac{\delta_2}{n_0}b_2\right) \\
        &\approx 1 + \frac{\delta_1 b_1 + \delta_2 b_2}{n_0}
    \end{align*}

    But, we know that this should be equivalent to the formulation where we consider a fixed utility of data throughout training, given by $b_\text{eff}$.
    \begin{align*}
        \frac{y_2}{y_0} &= \left(1 + \frac{\delta_1 + \delta_2}{n_0}\right)^{b_\text{eff}}
    \end{align*}

    Since the two formulations should be equivalent, we have that
    \begin{align*}
        b_\text{eff} = \frac{\delta_1 b_1 + \delta_2 b_2}{\delta_1 + \delta_2}
    \end{align*}

    Notice that in the final formulation, $b_\text{eff}$ is a weighted average of the utility of the data in the two buckets, with the weights being the effective data in the system at the time of training. However, this contradicts our initial assumption that it is only the data that decays in the system, and the utility stays constant. While the utility does not `decay', it does get re-weighted as the effective data in the system changes. This is a key insight that we gain from this analysis, which requires us to consider the change in both the effective data and the utility of data in the system. Note that these results still do not violate the formulation proposed in the work of~\citet{muennighoff2023scaling}, because they consider data to come from a single bucket. This means that $b_\text{eff} = \nicefrac{\delta_1 b_1}{\delta_1} = b_1$. However, in the case of a mixture of data buckets, the utility of data in the system is a weighted average of the utility of data in the individual buckets.

    An alternate paradigm in which such an equation may work is when the rate of decay ($\delta$) associated with all buckets is the same. However, we find that adding the constraint of fixed decay factor between all buckets leads to 
    significantly higher loss even on the points to be fit for scaling law, let alone the extrapolation points.

    \subsection{Decaying Effective Data with changing Utility}
    Based on the insights from the previous two formulations, we now consider a formulation that models the decay of both the utility of data and the effective data in the system. 
    It is clear from the previous part that we need a formulation where we can model the instantaneous utility of data in order to be able to mix different data buckets.
    The key difference from the formulation based on only utility decay is that we consider that there is a decay factor associated with the effective number of samples in the system. Let us first rewrite the instantaneous system state in terms of the effective data and utility of data as follows:
    \begin{equation}
    \begin{split} 
        \frac{dy}{dn_\text{eff}} &= \frac{y(n_\text{eff})}{n_\text{eff}}b,\\
        \frac{dn_\text{eff}}{dn} &= \delta^{k-1}
        \end{split} 
    \end{equation}

    From the formulation in the previous part in Section~\ref{app:subsubsec_effective_data_formulation}, it then follows that,
    \begin{align*}
        b_{\text{eff}}^{(k)} &= \frac{b_1 \delta_1^{k-1} + b_2 \delta_2^{k-1}}{\delta_1^{k-1} + \delta_2^{k-1}},
    \end{align*}
    where $b_{\text{eff}}^{(k)}$ is the effective utility of the mixture of data buckets at any given epoch $k$ of training. 

    Using the above, we can consider the closed form solution for the loss of the system at any given epoch $k$ of training as a
    product of the loss at the end of the first epoch, and the ratio of the effective data seen at the end of the $k$th epoch to the power effective utility.
    \begin{align*}
        y_k &=  y_1 \left(\frac{n_\text{eff2}}{n_\text{eff1}}\right)^{b_{\text{eff}}^{(2)}} \left(\frac{n_\text{eff3}}{n_\text{eff2}}\right)^{b_{\text{eff}}^{(3)}} \cdots \left(\frac{n_\text{effk}}{n_\text{eff({k-1})}}\right)^{b_{\text{eff}}^{(k)}}\\
        y_k &=  y_1 \prod_{j=2}^k \left(\frac{n_\text{effj}}{n_\text{eff(j-1)}}\right)^{b_{\text{eff}}^{(j)}}
    \end{align*}

    \subsubsection{Estimating Data Mixtures under Decaying Effective Data with changing Utility}
    We will directly use the data mixing results from \ref{app:subsubsec_effective_data_formulation} to estimate the loss of the system at any given epoch $k$ of training. Note that this is governed under the assumptions of exponential decay of data with repetitions.

    \subsection{Equivalence of Formulations F.1 and F.3}
    We will now show that under certain conditions, the decay of data and utility, and the decay of just utility reduce to the same formulation. Let us consider the case where the decay factor of the effective data is the same as the decay factor of the utility of data. Consider two buckets parametrized by $(b_1, \delta)$ and $(b_2, \delta)$ as their respective utility and decay factor. 
    We will now study the ratio of the two losses after seeing a single sample from the two buckets. Once again assume that the initial system state is characterized by $(y_0, n_0)$. For simplicity, we will consider the case of training on the second epoch of data:

    \paragraph{Effecttive Utility Formulation}
    Recall that 
    \begin{align*}
        b_\text{eff} = \frac{b_1 \delta^{k-1} + b_2 \delta^{k-1}}{2}.
    \end{align*}
    Then, we can write the loss after seeing a single sample as:
    \begin{align*}
    \frac{y_1}{y_0} &= \left(1 + \frac{1}{n_0}\right)^{b_\text{eff}} \\
    &= \left(1 + \frac{1}{n_0}\right)^{\frac{b_1 \delta_1 + b_2 \delta_2}{2}} \\
    &\approx \left(1 + \frac{1}{n_0}\right)^{\frac{b_1\delta_1 + b_2 \delta_2}{2}} \\
    &\approx \left(1 + \frac{b_1\delta_1 + b_2 \delta_2}{2n_0}\right)
    \end{align*}

    \paragraph{Effective Data, Dynamic Utility Based Formulation}
    In this case, the 
    \begin{align*}
        b_\text{eff} &= \frac{\delta_1 b_1 + \delta_2 b_2}{\delta_1 + \delta_2} \\
        \delta_\text{eff} &= \frac{\delta_1 + \delta_2}{2}
    \end{align*}

    Then, we can write the loss after seeing a single sample from the two buckets as:
    \begin{align*}
        \frac{y_1}{y_0} &= \left(1 + \frac{\delta_\text{eff}}{n_0}\right)^{b_\text{eff}} \\
        &= \left(1 + \frac{\delta_1 + \delta_2}{2n_0}\right)^{\frac{\delta_1 b_1 + \delta_2 b_2}{\delta_1 + \delta_2}} \\
        &\approx \left(1 + \frac{\delta_1 b_1 + \delta_2 b_2}{2n_0}\right) \\
    \end{align*}

    Notice that, under the Taylor approximation, and assumption that $n_0 \gg 1$, the two formulations are equivalent. This suggests that under the assumption that the decay factor of the effective data is the same as the decay factor of the utility of data, the two formulations reduce to the same formulation. This is a key insight that we gain from this analysis, which requires us to consider the change in both the effective data and the utility of data in the system. Note that these results still do not violate the formulation proposed in the work of~\citet{muennighoff2023scaling}, because they consider data to come from a single bucket. This means that $b_\text{eff} = \nicefrac{\delta_1 b_1}{\delta_1} = b_1$. However, in the case of a mixture of data buckets, the utility of data in the system is a weighted average of the utility of data in the individual buckets.

\section{Rate of Repetition Decay}
\label{app:sec:rate_repeat_decay}

Conventionally scaling laws have been studied in the context of language models~\citep{kaplan2020scaling,chinchilla}.
To the best of our knowledge, the only large scale attempt at predicting scaling of vision language models
was done in the work of~\citet{cherti2023reproducible}. However, the extrapolated graphs in their work
showed large errors as compared to the actual performance of the models. 
Hence, the scaling of vision language models is still an open question, especially when trained under contrastive loss.

\paragraph{Challenge: Squared Contrastive Pairs}
This is a fundamentally new challenge because in a data pool of $N$ samples, there exist $N^2$ pairs of comparisons that contribute to the loss. In such a setting, 
\emph{should the utility of a data pool still diminish exponentially after seeing $N$ samples, or should it diminish after seeing $N^2$ pairs?}
This is a fundamental question that we address in this section.

The contrastive loss for any given batch of data can be given by the following equation:
\begin{equation}
    \mathcal{L} = -\log \frac{\ell((x_i, t_i))}{\sum_{k=1}^{B-1} \ell(x_i, t_k)},
\end{equation}
where $B$ is the batch size, and $\ell$ is some loss function based on the cosine similarity between the embeddings of the image and the text.
$x_i$ and $t_i$ are the image and text embeddings of the $i$th sample in the batch. The goal is to maximize the similarity between the image and text embeddings of the same sample, and minimize the similarity between the embeddings of different samples.
We can decompose the loss into the numerator and the denominator. In a given batch, there are a total of $B$ samples seen in the numerator, and $B \times (B-1)$ samples seen in the denominator. We will 
call this $B \times B$ for simplicity.

Let us assume that the training set up uses a batch size of $B$ for contrastive learning. Given a dataset of $N$ samples, the number of batches seen in a single epoch is given by $\frac{N}{B}$.
Therefore, the total number of comparisions seen in the denominator in a single epoch is given by $N \times B$.

\paragraph{Utility Decay Assumption}
We assume that the utility decays exponentially, however, only after seeing each unique comparison in the dataset.
For a data pool of $N$ samples, while conventionally such a decay event would happen every epoch after seeing $N$ samples, in the case of contrastive learning, this decay event should happen after seeing $N^2$ pairs of samples.
This means that the number of epochs after which the utility of the data pool decays should be given by $\frac{N^2}{N \times B} = \frac{N}{B}$.

\paragraph{Decay formulation}
Let us assume that $\delta_\text{g}$ is the gold rate of decay after seeing $N^2$ samples. 
Since the modeling till now was done based on utility decay per epoch, we can write a relationship between the gold rate of decay and the rate of decay after seeing $N$ samples as follows:
\begin{align*}
    \delta_g &= \delta^{N/B} = \frac{1}{2}^{N/B\tau}
\end{align*}

\paragraph{Finding decay value on merging data buckets}
When we merge data buckets, a unique phenomenon happens. The overall pool size increases. While in the conventional language modeling
paradigm this may not have a significant impact, in the case of contrastive learning, this changes decay rate because the 
number of comparisons in the denominator increase at the squared rate of total samples.
Using $\delta_g$ we can find the new value of $\hat{\tau}$ as follows:
\begin{align*}
    \delta_g &= \frac{1}{2}^{N/B\tau} = \frac{1}{2}^{\hat{N}/B\hat{\tau}} \\
    \hat{\tau} &= \frac{\hat{N}}{N}\tau
\end{align*}

Hence, $\tau_p = p\tau$ when merging $p$ different buckets of a given size.

\begin{figure}
    \centering
    \includegraphics[width = 0.8\linewidth]{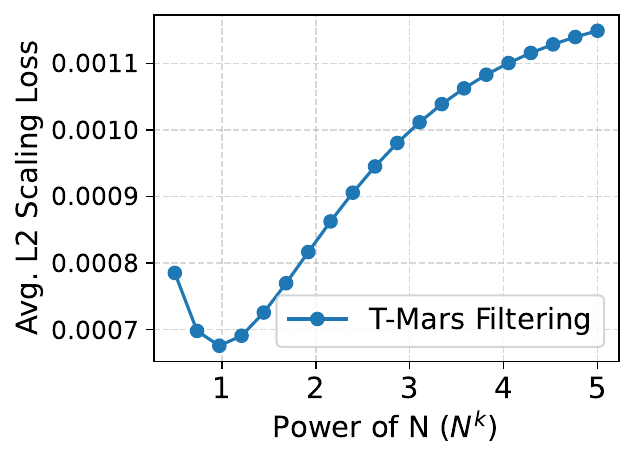}
    \caption{Loss surface as we find the best rate of decay of utility as a function of $N^k$.}
    \label{fig:finding_k_power}
\end{figure}

\paragraph{Finding the best decay rate}
In the above discussion, we assumed that the rate of decay should happen after seeing $N^2$ pairs of samples. However, this is an assumption that we made. 
To test how this empirically holds with respect to different rates of change in $N$ we allow for the flexibility of 
$\tau_p = p^k\tau$. Then we run an estimation analysis, at various values of $k$ as in Figure~\ref{fig:finding_k_power} to find the best value of $k$ that minimizes the loss on the data points to be fit.
We notice that the results of the theoretically derived relationship exactly match the empirically determined value of minimum loss at $k=1$.
Hence, for the purposes of all the merging experiments, we use $\tau_p = p\tau$ as the rate of decay of utility of data in the system when merging $p$ different data buckets.

\end{document}